\def\year{2020}\relax
\documentclass[letterpaper]{article} 
\usepackage{aaai20}  
\usepackage{times}  
\usepackage{helvet} 
\usepackage{courier}  
\usepackage[hyphens]{url}  
\usepackage{graphicx} 

\usepackage{amsfonts}
\usepackage{mathrsfs}
\usepackage{bbm}
\usepackage{footnote}
\usepackage{multirow}
\usepackage{amsmath}
\usepackage{amssymb}
\usepackage{mathrsfs}
\usepackage{bbm}
\usepackage{balance}
\usepackage{url}
\usepackage{graphicx}
\usepackage{color}
\usepackage{subfigure}
\usepackage{algorithm}
\usepackage{algorithmic}
\usepackage{paralist}
\usepackage{booktabs}
\usepackage{array}
\usepackage{csquotes}
\newcommand{\R}{\mathbb{R}}

\DeclareMathOperator*{\argmax}{arg\,max}

\usepackage{graphicx}  
\title{Learning Multi-level Dependencies for Robust Word Recognition}

\author{Zhiwei Wang\textsuperscript{\rm 1}\thanks{Work was done when interned at TAL AI Lab}, Hui Liu\textsuperscript{\rm 1}, Jiliang Tang\textsuperscript{\rm 1}, Songfan Yang\textsuperscript{\rm 2}, Gale Yan Huang\textsuperscript{\rm 2}, Zitao Liu\textsuperscript{\rm 2}\thanks{The corresponding author}\\ 
\textsuperscript{\rm 1}Michigan State University, \{wangzh65, liuhui7, tangjili\}@msu.edu\\ 
\textsuperscript{\rm 2}TAI AI Lab, TAL Education Group, \{yangsongfan, galehuang, liuzitao\}@100tal.com
}

\begin{document}

\maketitle

\begin{abstract}
Robust language processing systems are becoming increasingly important given the recent awareness of dangerous situations where brittle machine learning models can be easily broken with the presence of noises. In this paper, we introduce a robust word recognition framework that captures multi-level sequential dependencies in noised sentences.  The proposed framework employs a sequence-to-sequence model over characters of each word, whose output is given to a word-level bi-directional recurrent neural network. We conduct extensive experiments to verify the effectiveness of the framework. The results show that the proposed framework outperforms state-of-the-art methods by a large margin and they also suggest that character-level dependencies can play an important role in word recognition. The code of the proposed framework and the major experiments are publicly available\footnote{\url{https://github.com/DSE-MSU/MUDE}}.
\end{abstract}
\section{Introduction}
Most of the widely used language processing systems have been built on neural networks that are highly effective, achieving the performance comparable to humans~\cite{devlin2018bert,yang2019xlnet,yu2018qanet}. They are also very brittle, however, as they could be easily broken with the presence of noises~\cite{belinkov2017synthetic,zhao2017generating,ebrahimi2017hotflip}. However, the language processing mechanism of humans are very robust. One representative example is the following {\it Cambridge} sentence:

\begin{displayquote}
{\it Aoccdrnig to a rscheearch at Cmabrigde Uinervtisy, it deosn't mttaer in waht oredr the ltteers in a wrod are, the olny iprmoetnt tihng is taht the frist and lsat ltteer be at the rghit pclae. The rset can be a toatl mses and you can sitll raed it wouthit porbelm. Tihs is bcuseae the huamn mnid deos not raed ervey lteter by istlef, but the wrod as a wlohe.}
\end{displayquote}

In spite of the fact that a human can read the above sentence with little difficulty, it can cause a complete failure to existing natural language processing systems such as Google Translation Engine~\footnote{https://translate.google.com/}. Building robust natural language processing systems is becoming increasingly important nowadays given severe consequences that can be made by adversarial samples~\cite{xu2019adversarial}: carefully misspelled spam emails that fool spam detection systems~\cite{fumera2006spam} deliberately designed input sentences that force chatbot to emit offensive language~\cite{wolf2017we,dinan2019build,liu2019say}, etc. Thus, in this work, we focus on building a word recognition framework which can denoise the  misspellings such as those shown in the {\it Cambridge} sentence. As suggested by psycholinguistic studies~\cite{rayner2006raeding,davis2012psycholinguistic}, the humans can comprehend text that is noised by jumbling internal characters while leaving the first and last characters of a word unchanged. Thus, an ideal word recognition model is expected to emulate robustness of human language processing mechanism. 

The benefits of such framework are two-folds. The first is its recognition ability can be straightforwardly used to correct misspellings. The second is its contribution to the robustness of other natural language processing systems. By serving as a denoising component, the word recognition framework can firstly clean the noised sentences before they are inputted into other natural language processing systems~\cite{pruthi2019combating,zhou2019improving}.

\begin{figure*}[h!]
    \begin{center}
        \subfigure[Train Stage.]{\label{fig:train}\includegraphics[scale=0.51]{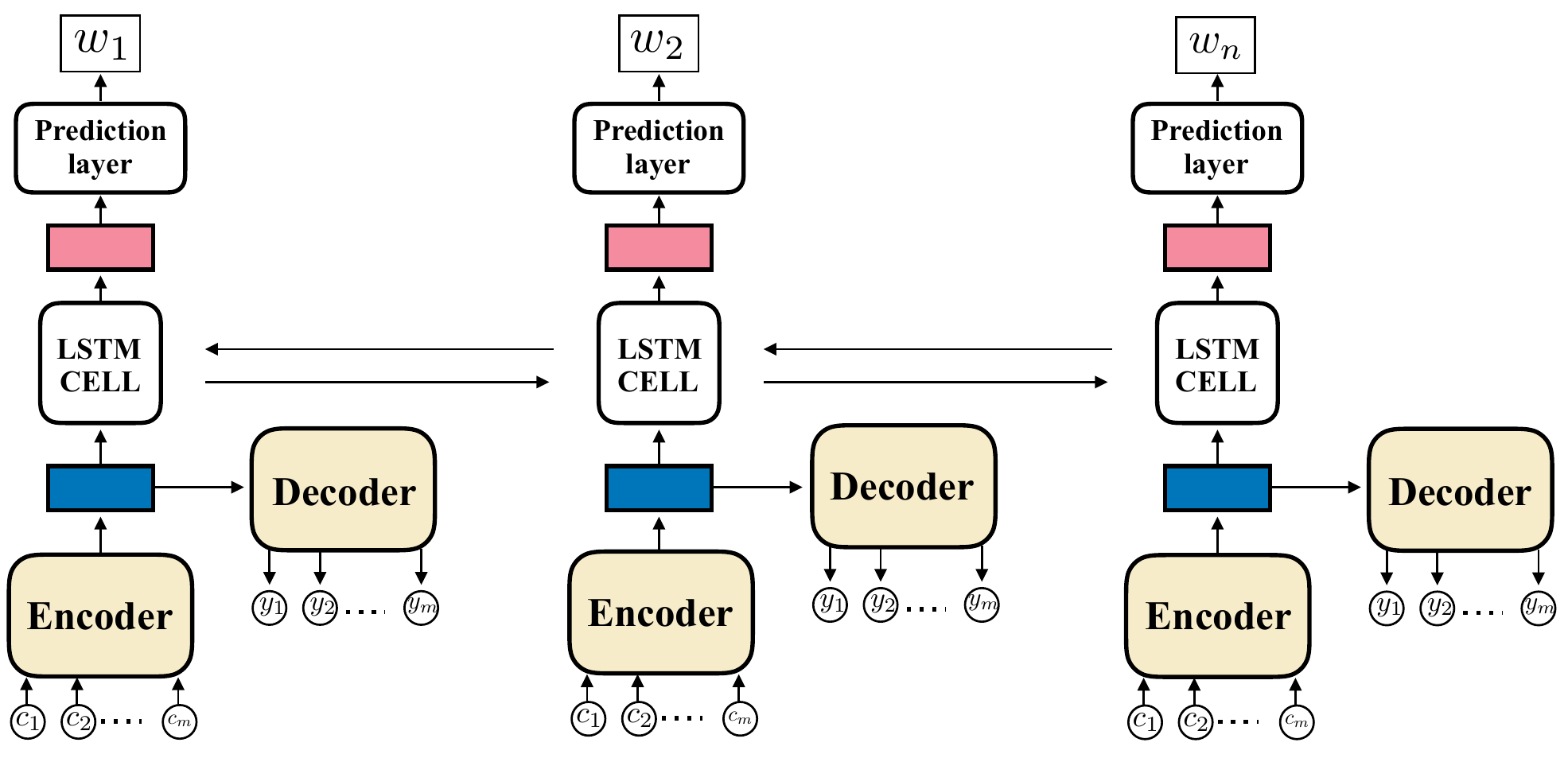}}\hspace{19mm}
        \subfigure[Test Stage]{\label{fig:test}\includegraphics[scale=0.51]{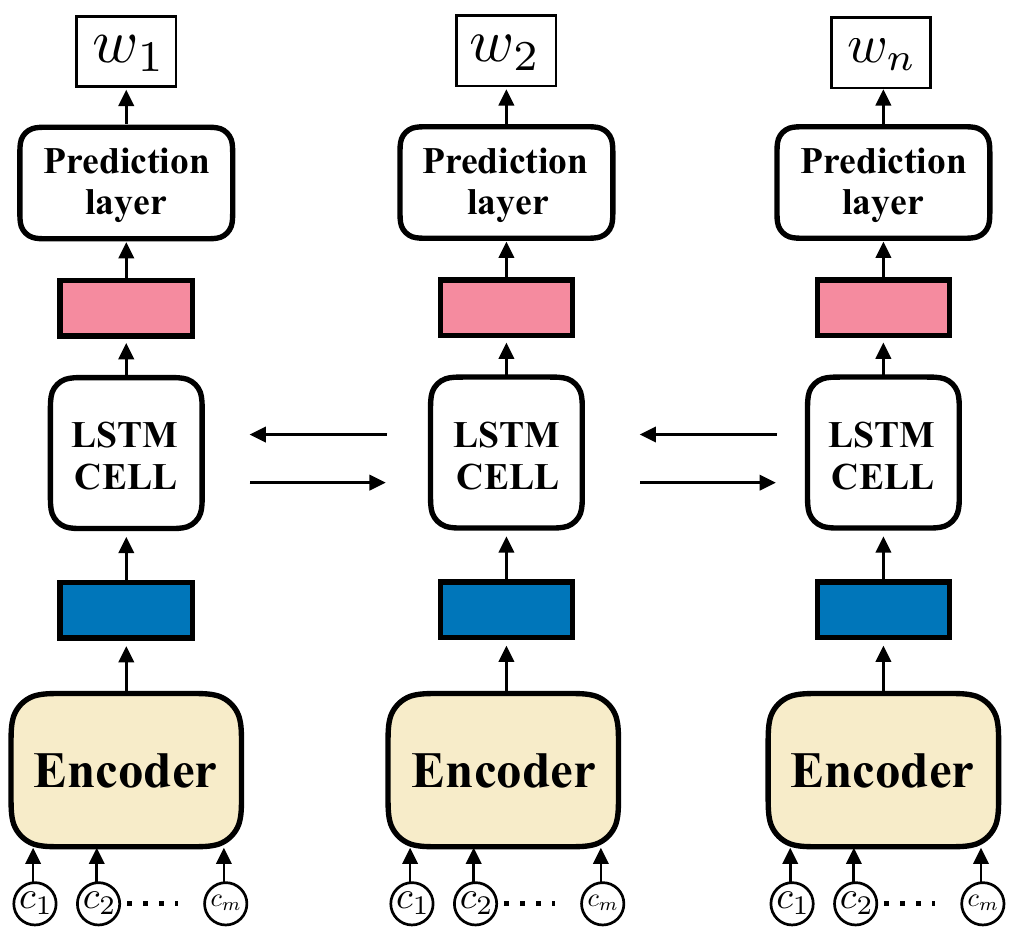}}
    \end{center}
    \caption{The graphical illustration of the proposed framework: MUDE.}
    \label{fig:model}
\end{figure*}

From the human perspective, there are two types of information that play an essential role for us to recognize the noised words~\cite{perea2015letter}.  The first is the character-level dependencies. Take the word `{\it wlohe}' in the {\it Cambridge} sentences as an example, it is extremely rare to see a `w' sits next to an `l' in an  English word. Instead, it is more natural with `wh'. Thus, it is quite easy for humans to narrow down possible correct forms of `{\it wlohe}' to be `{\it whole}' or `{\it whelo}'. To ensure that it should be `{\it whole}', we often need the second type of information: context information such as `{\it but the wrod as a wlohe.}', which is denoted as word-level dependencies in this paper. Intuitively, an effective word recognition framework should capture these multi-level dependencies. However, multi-level dependencies are rarely exploited by the exiting works such as scRNN~\cite{sakaguchi2017robsut}. Hence, we propose a framework MUDE that is able to fully utilize {\bf mu}lti-level {\bf de}pendencies for the robust word recognition task. It integrates a character-level sequence-to-sequence model and a word-level sequential learning model into a coherent framework. The major contributions of our work are summarized as follows:
\begin{itemize}
  \item We identify importance of character-level dependencies for recognizing a noised word;
  \item We propose a novel framework, MUDE, that utilizes both character-level and word-level dependencies for robust word recognition task;
  \item We conduct extensive experiments on various types of noises to verify the effectiveness of MUDE.
\end{itemize}

For the rest of the paper, we firstly give a detailed description of MUDE. Then we conduct experiments to verify its effectiveness. We will also show that MUDE is able to achieve the state-of-the-art performance no matter what type of noise presents and outperforms the widely used baselines by a large margin. Next, we introduce the relevant literature in the related work section, followed by a conclusion of current work and discussion of possible future research directions.


\section{The Proposed Framework: MUDE}
In this section, we describe MUDE that is able to capture both character-level and word-level dependencies. The overall architecture is illustrated in Figure~\ref{fig:model}. It consists of 3 major components: a sequence-to-sequence model, a bi-directional recurrent neural network and a prediction layer. Next we will detail each component.

\subsection{Learning Character-level Dependencies}
As mentioned previously, there exist sequential patterns in the characters of a word. For example, vocabulary roots such as {\it cur} and {\it bio} can be found in many words. In this subsection, we propose a sequence-to-sequence model to learn a better representation of a given word by incorporating character-level dependencies. The model consists of an encoder and a decoder, which we will describe next.

\subsubsection{Encoder}
Let ${\hat w}={c_1, c_2, \cdots c_m}$ be a sequence of characters of a given noised word ${\hat w}$.  We firstly map each character $c_i$ to a $d_c$-dimensional character embedding as follows:
\begin{align} 
x_i = {\bf E}o_i
\end{align}
\noindent where ${\bf E} \in \R^{C\times d_c}$ is the embedding matrix given that the total number of unique characters is $C$.  $o_i \in \R^{C}$ is the one-hot representation of $c_i$. Since there could some noise in ${\hat w}$, the sequential order of $c_i$ can be misleading. Thus, instead of using a sequential learning model such as recurrent neural network, we choose the multi-head attention mechanism~\cite{vaswani2017attention} to model the dependencies between characters without considering their order. To do so, we add a special character $c_0$ whose final representation will be used as the representation of the word. 

Specifically, the multi-head attention mechanism will obtain a refined representation for each character in ${\hat w}$. Next, without the loss of generality, we will use $c_i$ as an example to illustrate. To obtain the refined representation of $c_i$, $x_i$ will firstly be projected into query space and $x_j \: \forall j \in \{0,1,\cdots, m\}$ will be projected into key and value spaces as follows:
\begin{align}
\label{eq:projection}
x^q_i &= {\bf Q}x_i  \nonumber \\
x^k_j &= {\bf K}x_j \quad \forall j \in \{0,1,\cdots, m\}  \\
x^v_j &= {\bf V}x_j \quad \forall j \in \{0,1,\cdots, m\} \nonumber
\end{align}
\noindent where ${\bf Q}$, ${\bf K}$, ${\bf V}$ are the projection matrices for query, key, and value spaces, respectively. With $x^q_i$, $x^k_j$ and $x^v_j$, the refined representation $e_i$ of $c_i$ can be calculated as the weighted sum of $x^v_j$:
\begin{align}
\label{eq:att}
e_i = \sum \alpha_j x^v_j
\end{align}
\noindent where $\alpha_j$ is the attention score that is obtained by the following equation:
\begin{align}
\alpha_0, \alpha_1, \cdots, \alpha_m = softmax(\frac{{x^q_i}^Tx^k_0}{\sqrt{d}}, \frac{{x^q_i}^Tx^k_1}{\sqrt{d}}, \cdots, \frac{{x^q_i}^Tx^k_m}{\sqrt{d}})
\end{align}
To capture the dependencies of characters from different aspects, multiple sets of projection matrices are usually used, which will result in multiple sets of $x^q_i$, $x^k_j$ and $x^v_j$, and thus $e^i$. To be concrete, assume that there are $h$ sets of projection matrices, from Eq.~(\ref{eq:projection}) and Eq.~(\ref{eq:att}), we can obtain $h$ $e_i$s, which are denoted as \{$e^1_i, e^2_i, \cdots, e^h_i$\}. With this, the refined representation of $c_i$ is obtained by the concatenation operation:
\begin{align}
z_i = concatenation(e^1_i, e^2_i, \cdots, e^h_i)
\end{align}
\noindent where $z_i$ is the new representation of $c_i$ and contains dependency information of $c_i$ to other characters in ${\hat w}$ from $h$ aspects.

Following~\cite{vaswani2017attention}, we also add a positional-wise feedforward layer to $z_i$ as follows:
\begin{align}
p_i = {\bf W}^2ReLU({\bf W}^1z_i)
\end{align}
\noindent where ${\bf W}^1$ and ${\bf W}^2$ are the learnable parameters. $p_i$ is the final representation of $c_i$. Note that we can have several above mentioned layers stacked together to form a deep structure. 

At this point, we have obtained the refined representation vector for each character and we use that of the special character $c_0$ as the representation of given noised word, which is denoted as $w^c$

\subsubsection{Decoder} To capture the sequential dependency in the correct words, the Gated Recurrent Unit (GRU) which has achieved great performance in many sequence learning tasks~\cite{xu2019spatio,andermatt2016multi,xu2019adaptive} is used as the decoder. To be specific, in the decoding process, the initial hidden state $h_0$ of GRU is initialized with the noised word presentation ${\hat w}$. Then at each time stamp $t$, GRU will recursively output a hidden state $h_t$ given the hidden state $h_{t-1}$ at the previous time stamp. Due to the page limitation, we do not show the details of GRU, which is well described in~\cite{cho2014learning}. In addition, each hidden state will emit a predicted character $c^p_t$. The decoding process will end when the special character denoting the end of word is emitted. Concretely, the whole decoding process is formally formulated as follows:
\begin{align}
h_0 &= w^c \\ \nonumber
h_t &= GRU(h_{t-1}) \\ \nonumber
p_t &= softmax({\bf W}^p h_t) \\ \nonumber
c^p_t &= \argmax_i(p_t[i])
\end{align}
\noindent where ${\bf W}^p \in \R^{C\times d}$ is a trainable parameter. $p_t \in \R^{C}$ gives the emission probability of each character and $p_t[i]$ denotes the $i^{th}$ entry of vector $p_t$.

\subsubsection{Sequence-to-sequence Loss}
To train the previously described character-level sequence-to-sequence model, we define the loss function as follows:
\begin{align}
\mathcal{L}_{seq2seq} = -\sum_i^m p_i[y_i]
\end{align}
\noindent where $y_i$ is the index of the ground truth at position $i$ of the correct word $w$. By minimizing $\mathcal{L}_{seq2seq}$, the designed sequence-to-sequence model can learn a meaningful representation that incorporates character-level sequential dependencies for the noised word. Next, we will describe the framework component that captures the word-level dependencies. 

\subsection{Capturing Word-level Dependencies}
From the human perspective, it is vitally important to consider the context of the whole sentences in order to understand a noised word. For example, it would be very hard to know `frist' means `first' until a context `the olny iprmoetnt tihng is taht the frist and lsat ltteer be at the rghit pclae.' is given. Thus, to utilize the context information and word-level dependencies, we design a recurrent neural network (RNN) to incorporate them in the noised word representation. Specifically, the word presentations obtained from character-level encoder will be passed into a bi-directional long short-term memory (LSTM). Concretely, given a sequence of word presentations $S=\{w^c_1, w^c_2, \cdots, w^c_n\}$ obtained from character-level dependencies, we calculate a sequence of refined word representation vectors $\{w_1, w_2, \cdots, w_n\}$ as follows:
\begin{align}
w^f_1, w^f_2, \cdots, w^f_n &= LSTM_{forward}(w^c_1, w^c_2, \cdots, w^c_n) \nonumber \\
w^b_1, w^b_2, \cdots, w^b_n &= LSTM_{backward}(w^c_1, w^c_2, \cdots, w^c_n)  \\ 
w_1, w_2, \cdots, w_n &= w^f_1||w^b_1, w^f_2||w^b_2, \cdots, w^f_n||w^b_n \nonumber
\end{align}
\noindent where $\|$ denotes concatenation. $LSTM_{forward}$ indicates that $w^c$s are processed from $w^c_1$ to $w^c_n$, while $LSTM_{backward}$ processes word presentations in an opposite direction, namely, from $w^c_n$ to $w^c_1$.  Comparing to original LSTM where only forward pass is performed, bi-directional LSTM can include both `past' and `future' information in the representation of $w_i$. 

With the aforementioned procedure, the representation of each word now incorporates both character-level and word-level dependencies. Thus, the correct word is predicted as follows:
\begin{align}
p^w_i &= softmax({\bf W}^w w_i) \\ \nonumber
w^p_i &= \argmax_i(p^w_t[i]) 
\end{align}
\noindent where ${\bf W}^w \in \R^{V\times d_w}$ is a trainable matrix and $V$ is the size of the vocabulary that contains all possible words. Moreover, $p^w_i \in \R^{V}$ is the probability distribution over the vocabulary for the $i^{th}$ word in a sentence. 

\subsubsection{Word Prediction Loss} To effectively train MUDE for correct word prediction, similar to character-level sequence-to-sequence model, we define the following objective function: 

\begin{align}
\mathcal{L}_{pred} =  -\sum_i^n p^w_i[y^w_i]
\end{align}
\noindent where $y^w_i$ is the index of the $i^{th}$ correct word.

\subsection{Training Procedure}
So far we have described MUDE which includes a character-level sequence-to-sequence model and a word-level sequential learning model. To train both models simultaneously, we design a loss function for the whole framework as follows:
\begin{align}
\label{eq:loss}
\mathcal{L} = \mathcal{L}_{pred} + \beta \mathcal{L}_{seq2seq}
\end{align}
\noindent where $\beta$ is a hyperparameter that controls the contribution of the character-level sequence-to-sequence model. Since the major goal of the framework is to predict the correct word given the noised word, we decrease the value of $\beta$ gradually as the training proceeds to allow the optimizer increasingly focus on improving the word prediction performance.

\subsubsection{Test Stage}
As shown in Figure~\ref{fig:model}, in the test stage, we simply remove the decoder of the sequence-to-sequence model and only keep the encoder in the framework.

\section{Experiment}
In this section, we conduct extensive experiments on the spell correction task to verify the effectiveness of MUDE. Next, we firstly introduce the experimental settings, followed by the analysis of the experimental results.
\subsection{Experimental Settings}

\subsubsection{Data}
We use the publicly available Penn Treebank~\cite{marcus1993building} as the dataset. Following the previous work~\cite{sakaguchi2017robsut}, we firstly experiment on 4 different types of noise: Permutation (PER), Deletion (DEL), Insertion (DEL), and Substitution (SUB), which only operate on the internal characters of words, leaving the first and last characters unchanged. Table~\ref{tab:noise_example} shows a toy example of a noised sentence. These 4 types of noise can cover most of the realistic cases of misspellings and commonly tested in previous works~\cite{belinkov2017synthetic,pruthi2019combating}. 
For each type of noise, we construct a noised dataset from the original dataset by altering all the words that have more than 3 characters with corresponding noise. We use the same training, validation and testing split in~\cite{sakaguchi2017robsut}, which contains 39,832, 1,700 and 2,416 sentences, respectively. 
\begin{table}[h!]
    \begin{center}
  \caption{Toy examples of noised text}
  \begin{tabular}{lc}
  \toprule \\ [-1.7ex]
    Noise Type & Sentence \\
    \\ [-1.7ex]
    \hline
    \\ [-1.7ex]
    Correct & An illustrative example of noised text\\
    PER & An isulvtriatle epaxmle of nsieod txet\\
    DEL & An illstrative examle of nosed tet \\
    INS & An ilelustrative edxample of nmoised texut \\
    SUB & An ilkustrative exsmple of npised test \\
    \\ [-1.7ex]
 \bottomrule 
  \end{tabular}
  \label{tab:noise_example}
  \end{center}
\end{table}

\subsubsection{Baselines}
To show the effectiveness of MUDE, we compare it with two strong and widely used baselines. The first is Enchant~\footnote{https://abiword.github.io/enchant/} spell checker which is based on dictionaries. The second one is scRNN~\cite{sakaguchi2017robsut}. It is a recurrent neural network based word recognition model and has achieved previous state-of-the-art results on spell correction tasks. This baseline only considers the sequential dependencies in the word level with a recurrent neural network and ignores that of character level. Note that other baselines including CharCNN~\cite{sutskever2011generating} have been significantly outperformed by scRNN. Thus, we do not include them in the experiments.

\subsubsection{Implementation Details}

Both scRNN and MUDE are implemented with Pytorch. The number of hidden units of word representations is set to be 650 as suggested by previous work~\cite{sakaguchi2017robsut}. The learning rate is chosen from \{0.1, 0.01, 0.001, 0.0001\} and $\beta$ in Eq~(\ref{eq:loss}) is chosen from \{1, 0.1, 0.001\} according to the model performance on the validation datasets. The parameters of MUDE are learned with stochastic gradient decent algorithm and we choose RMSprop~\cite{tieleman2012rmsprop} to be the optimizer as it did in~\cite{sakaguchi2017robsut}. To make the comparison fair, scRNN is trained with the same settings as MUDE.

\subsection{Comparison Results}

The comparison results are shown in Table~\ref{tab:result_1}. There are several observations can be made from the table. The first is that model based methods (scRNN and MUDE) achieve much better performance than dictionary based one (Enchant). This is not surprising as model based methods can effectively utilize the sequential dependencies of words in the sentences. Moreover, MUDE consistently outperforms scRNN in all cases, which we believe attributes to the effective design of MUDE to capture both character and word level dependencies. More detailed analysis of contribution brought by the character-level dependencies will be shown later in this section. In addition, we observe that the difficulty brought by different types of noise varies significantly. Generally, for model based methods, permutation and insertion noises are relatively easier to deal with comparing to deletion and substitution noises. We argue this is because the former ones do not lose any character information. In other words, the original character information is largely preserved with permutation and insertion. On the contrary, both deletion and substitution can cause information loss, which makes it harder to recognize the original words. This again demonstrate how important the character-level information is. Finally, the results also show that in more difficult situations where deletion or substitution noises present, the advantages of the MUDE become even more obvious. This clearly suggests the effectiveness of the MUDE. 

\begin{table}[h!]
  \begin{center}
  \caption{Performance comparison with different types of noise in terms of accuracy (\%). Best results are highlighted with bold numbers.}
  \vspace{2mm}
  \begin{tabular}{lcccc}
  \toprule 
    Method  & INT & DEL  & INS & SUB \\
    \hline \\ [-1.7ex]
    Enchant & 72.33  & 71.23 & 93.93  & 79.77\\
    scRNN & 98.23 &  91.55 & 95.95  & 87.09\\
    MUDE & {\bf 98.81} & {\bf 95.86} & {\bf 97.16}  & {\bf 90.52}\\
 \bottomrule 
  \end{tabular}
  \label{tab:result_1}
  \end{center}
\end{table}

Next, we take one step further by removing the constraint that the noise will not affect the first and last characters of each word. More specifically, we define 4 new types of noise that are W-PER, W-DEL, W-INS, and W-SUB, which stand for altering a word by permuting the whole word, deleting, inserting, and substituting characters in any position of the word. Similarly, for each type of new noise, we construct a noised dataset. The results are shown in Table~\ref{tab:result_2}. 
\begin{figure}[h!]
    \begin{center}
        \subfigure[Prediction loss.]{\label{fig:pred_loss}\includegraphics[scale=0.38]{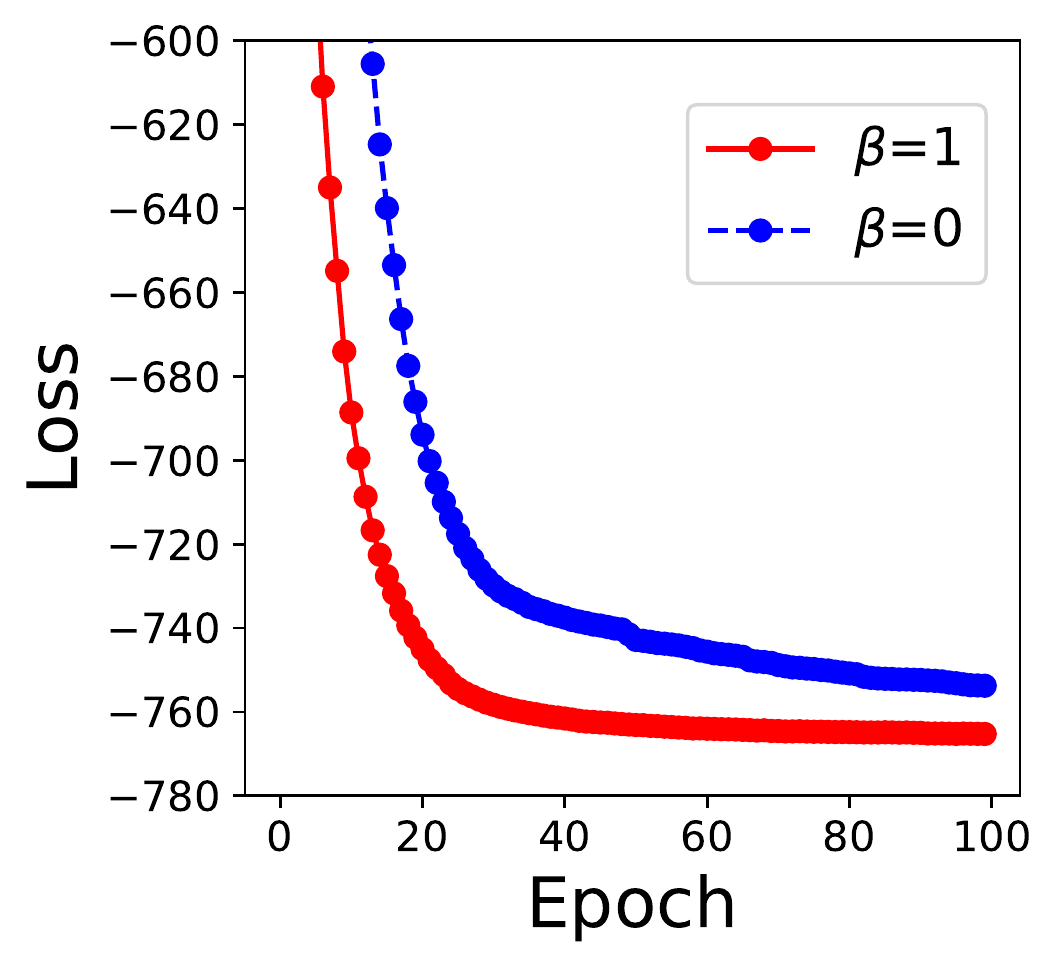}}
        \subfigure[Seq2Seq loss.]{\label{fig:seq_loss}\includegraphics[scale=0.38]{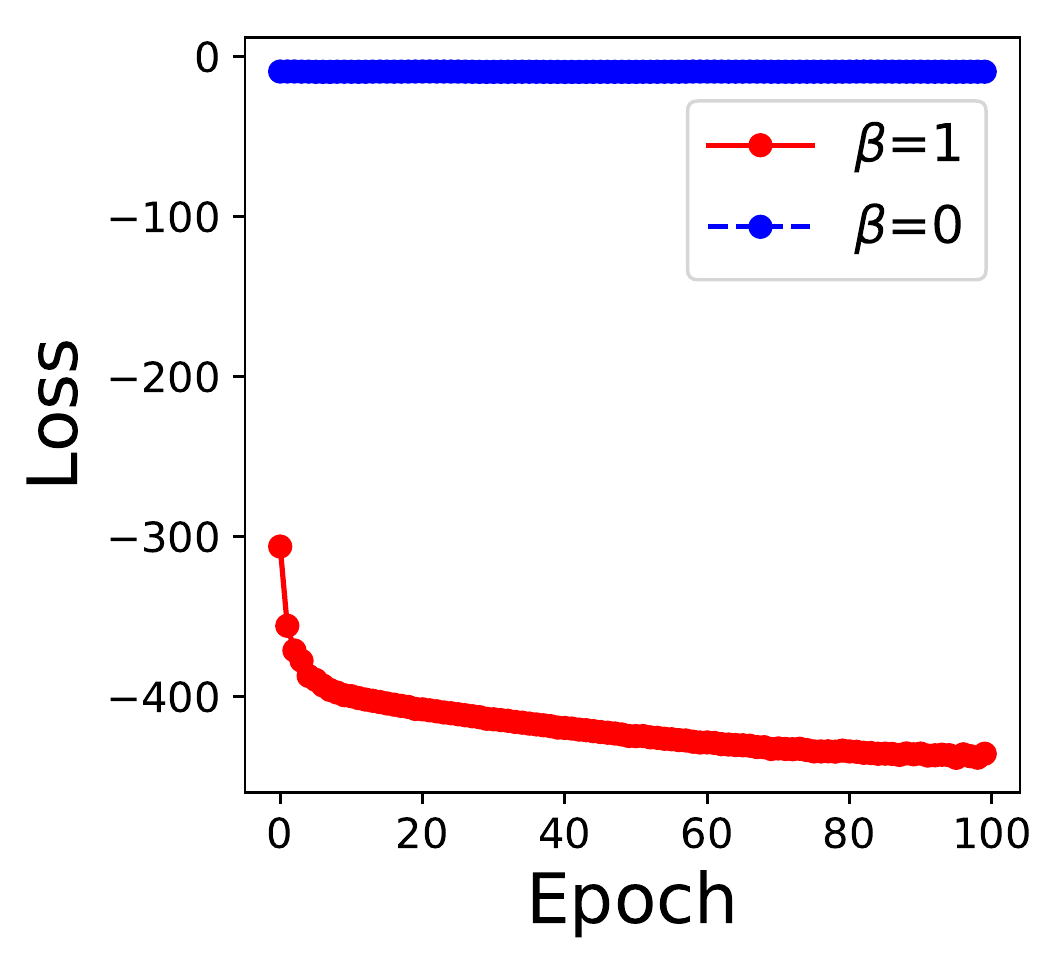}}
    \end{center}
    \caption{Learning curve of MUDE in the training procedure with different $\beta$ values.}
    \label{fig:para}
\end{figure}
\begin{table}
    \begin{center}
  \caption{Performance comparison different type of noise in terms of accuracy (\%). Best results are highlighted with bold numbers.}
  \vspace{2mm}
  \begin{tabular}{lcccccc}
  \toprule 
    Method  & W-PER &  W-DEL & W-INS & W-SUB\\
    \hline \\ [-1.7ex]
    Enchant  & 59.08 & 69.84  & 93.77 & 77.23\\
    scRNN  & 97.36  & 89.99  & 95.96 & 81.12\\
    MUDE  & {\bf 98.75}  & {\bf 93.29} & {\bf 97.10} & {\bf 85.17} \\
 \bottomrule 
  \end{tabular}
  \label{tab:result_2}
  \end{center}
\end{table}

From the table, we observe that firstly, the performance of nearly all methods decreases comparing to that of Table~\ref{tab:result_1}. This suggests the new types of noise are more difficult to handle, which is expected as they cause more variations of noised words. In fact, without keeping first and last characters of each words, it also becomes a difficult task for human to comprehend the noised sentences~\cite{rayner2006raeding}. Secondly, MUDE still achieves higher accuracy than other baselines, which is consistent with observations from Table~\ref{tab:result_1}. More importantly, as the difficulty of the task increases, the advantages of MUDE over scRNN also become more obvious. Take the noise of substitution for example, in Table~\ref{tab:result_1}, MUDE has around 3.5\% absolute accuracy gain over scRNN. When more difficult noise (W-SUB) comes, the performance gain of MUDE becomes 4\% as shown in Table~\ref{tab:result_2}. Such observation is also consistent with previous findings. 

In summary, both Table~\ref{tab:result_1} and ~\ref{tab:result_2} clearly demonstrate the robustness of MUDE and its advantages over scRNN which can not utilize the character-level dependencies. Thus, in the next subsection, we conduct analysis on the contribution of character-level dependencies to gain better understanding of MUDE.

\subsection{Parameter Analysis}

In this subsection, we analyze the contribution of character-level dependencies to better word representations by showing the model performance with different $\beta$ values, which controls the contribution of character-level sequence-to-sequence loss. Specifically, we let the $\beta$ be $0$ and $1$. When $\beta$ is $0$, MUDE will totally ignore the character-level dependencies; When $\beta$ equals to $1$, MUDE achieve best accuracy in validation set. The prediction loss and seq2seq loss during the training stage with different $\beta$ values are shown in Fig~\ref{fig:para}. Note that the trends in Fig~\ref{fig:para} are similar in all of the cases with the different types noise and we only show that of W-PER case due to the page limitation.

As the upper sub-figure shows, when $\beta=1$ the prediction loss converges faster and at a lower value comparing to that of case when $\beta=0$. For the seq2seq loss, it remains constant value when $\beta=0$ as the model does not learn anything regarding seq2seq task. On the other hand, when $\beta=1$, the seq2seq loss stably decreases, suggesting that the MUDE is trained to obtain better representation of each word. The obvious positive correlation between these two losses clearly demonstrates the importance of learning character-level dependencies in misspelling correction tasks.

\begin{table*}[h!]
    \begin{center}
  \caption{Generalization analysis results. The best result are highlighted. MEAN shows the average value of each row.}
  \vspace{2mm}
  \begin{tabular}{clccccccccc}
  \toprule \\ [-1.7ex]
     &   & \multicolumn{8}{c}{Test Noise}& \\  \\ [-1.7ex]
    &  & PER & W-PER & DEL & W-DEL & INS & W-INS & SUB & W-SUB & MEAN\\
    \\[-1.7ex]
    \hline \\ [-1.7ex]
     \multirow{8}{*}{\shortstack[l]{Train \\ \\ Noise}}
    & PER  & -- & {\bf 98.81} & 82.55 & 79.61 & 92.21 & 92.37 & 71.39 & 69.88 & 85.70 \\
    & W-PER & {\bf 98.75} & -- &  81.31  & 78.3  & 91.32  & 91.25  & 69.55  & 67.91 & 84.64\\
    & DEL   & 90.83 & 90.83 & -- & {\bf 86.02} & 79.96 & 79.97 & 81.99 & 76.02 & 85.18\\
    & W-DEL & 86.75 & 86.75  & {\bf 94.08}  & -- & 78.83 & 78.87 & 80.35 & 79.07 & 84.74\\
    & INS   & 94.79 & 94.79 & 77.3  & 74.81 & -- & {\bf 97.15} & 82.86 & 80.42 & 87.41\\
    & W-INS & 95.67 & 95.67 & 78.34 & 75.95 & {\bf 97.01} & -- & 82.96 & 80.78  & {\bf 87.91}\\
    & SUB   & 91.71 & 91.71 & 88.34 & 81.49 & 81.19 & 81.21 & -- & {\bf 83.65} & 86.22 \\
    & W-SUB & 87.05 & 87.05 & 83.42 & 82.42 & 79.27 & 79.17 & {\bf 85.67} & -- & 83.65 \\

    \\ [-2.5ex]
 \bottomrule
  \end{tabular}
  \label{tab:result_3}
  \end{center}
\end{table*}

\begin{table*}[h!]
    \begin{center}
  \caption{Data augmentation results. The values that are higher than these of Table~\ref{tab:result_3} are bold.}
  \begin{tabular}{clccccccccc}
  \toprule \\ [-1.7ex]
     &   & \multicolumn{8}{c}{Test Noise}& \\  \\ [-1.7ex]
    &  & Per & W-PER & DEL & W-DEL & INS & W-INS & SUB & W-SUB & MEAN\\
    \\[-1.7ex]
    Train Noise & W-ALL   & 96.45 & 96.45 & {\bf 94.26} & {\bf 93.34} & 95.3 &  95.28 & {\bf 91.51} & {\bf 90.48} & {\bf 94.13}\\

    \\ [-2.5ex]
 \bottomrule
  \end{tabular}
  \label{tab:result_4}
  \end{center}
\end{table*}

\subsection{Generalization Analysis}
In this subsection, we conduct experiments to understand generalization ability of MUDE. Concretely, we train the framework on one type of noise and test it with a dataset that presents another type of noise. The results are shown in Table~\ref{tab:result_3}.

From results, we have the following two observations. Firstly, between datasets with similar type of noise, MUDE generalizes quite well (e.g. trained on W-PER and tested on PER), which is not surprising. However, the MUDE trained on one type of noise performs much worse on other types of noise that are very different.  These observations suggest that it is hard for MUDE to generalize between noises, which we argue is possibly because of the small overlap between distributions of each type of noise. 

Thus, in the next, we apply the commonly used adversarial training method by augmenting all types of noise to train MUDE and test it on each type noise individually. As W-* (*$\in $ \{PER, DEL, INS, SUB\}) includes the *, in this experiment, we only combine W-* instead of all types of noise. We denote the new constructed training dataset as W-ALL. The results are shown in Table~\ref{tab:result_4}. It can be observed from table that the MUDE trained on W-ALL has much better generalization ability (i.e., the mean value is much higher). In addition, it is interesting to see that performance of the MUDE decreases slightly in relatively easy cases where permutation or insertion noise presents while increasing a lot in difficult cases where deletion or substitution noise presents.

\subsection{Case Study}
In this subsection, we take the {\it Cambridge} sentences which are not the training set as an example to give a qualitative illustration of MUDE's misspelling correction performance. Note that due to the constraint of space, we only show the results of the two types of noise:  W-PER and W-INS. The example is shown in Table~\ref{tab:cambridge}. We can see from the table that it is quite difficult for even humans to comprehend the noised sentence when first and last characters are also changed. However, MUDE can still recognize almost all of the words. In addition, for both cases, the MUDE has much less errors in the corrected sentence than scRNN, which is consistent with previous quantitative results. 

\begin{table*}[h!]
\centering
\small.
\begin{tabular}{p{1.5cm}  p{13cm}} 
\toprule
Correct & According to a researcher at Cambridge University , it does n't matter in what order the letters in a word are , the only important thing is that the first and last letter be at the right place . The rest can be a total mess and you can still read it without problem . This is because the human mind does not read every letter by itself ,  but the word as a whole .

\\[1mm] \hline \\[-1mm]
\multicolumn{2}{c}{\bf W-PER} \\  [1mm] 
\hline \\[-1.5mm]
Noised  & iodrcAngc ot a reeachsr at meigaCdbr srtiinUyve , it seod tn' amrtte in wtah rerdo het tserelt in a rdwo rae , the onyl onmtiaptr ingth si tath hte itfrs dan stla treelt be ta het tgrhi place . hTe rset nca be a aotlt mess dan ouy anc lsilt drae ti tthwuoi lorbmpe . hTsi is aubeecs the huamn dmni edos nto erad evrye lteter by etfisl , but het rdwo sa a eholw .
\\[1mm] \hline \\[-1.5mm]
scRNN  & According to a \underline{\bf  research} at Cambridge University , it does n't matter in what order the letters in a word are , the only important thing is that the first and last letter be at the right place . The rest can be a total mess and you can still read it without problem . This is because the human mind does not read \underline{\bf  very} letter by itself , but the word as a whole .
\\[1mm] \hline \\[-1.5mm]
MUDE & According to a \underline{\bf  research} at Cambridge University , it does n't matter in what order the letters in a word are , the only important thing is that the first and last letter be at the right place . The rest can be a total mess and you can still read it without problem . This is because the human mind does not read every letter by itself , but the word as a whole .

\\[1mm] \hline \\[-1mm]
\multicolumn{2}{c}{\bf W-INS} \\  [2mm] 
\hline \\[-1.5mm]
Noised  & Acxcording to a reysearch at Cazmbridge Univversity , it doesw n't msatter in whmat orderh the letteros in a fword are , the oynly wimportant tghing is tyhat the fircst and ldast legtter be at the rightv placeu . The resty can be a totalp mesus and you can stillb rnead it withougt promblem . Txhis is bebcause the humgan minnd doess not reabd everyb lettfer by itslelf , but the whord as a whvole .
\\[1mm] \hline \\[-1.5mm]
scRNN  & according to a \underline{\bf  research} at Cambridge University , it does n't matter in what order the letters in a word are , the only important thing is that the first and last \underline{\bf  better} be at the right place . The rest can be a total \underline{\bf  less} and you can still read it without problem . This is because the human mind does not \underline{\bf  rated} every \underline{\bf  better} by itself , but the word \underline{\bf  a} a whole .
\\[1mm] \hline \\[-1.5mm]
MUDE & According to a \underline{\bf  research} at Cambridge University , it does n't matter in what order the letters in a word are , the only important thing is that the first and last letter be at the right place . The rest can be a total \underline{\bf  uses} and you can still read it without problem . This is because the human mind does not \underline{\bf  bear} every letter by itself , but the word as a whole .

\\ \bottomrule
\end{tabular}
\caption{An illustrative example of spelling correction outputs for the {\it Cambridge} sentence. Words that the models fail to correct are underlined and bold.}
\label{tab:cambridge}
\end{table*}

\section{Related Work}
In this section, we briefly review the related literature that is grouped into two categories. The first category includes the exiting works on similar tasks and the second one contains previous works that have applied word recognition model to improve the robustness of other NLP systems.
\subsection{Grammatical Error Correction}
Since the CoNLL-2014 shared task~\cite{ng2014conll}, Grammatical Error Correction (GEC) has gained great attention from NLP communities~\cite{zhao2019improving,grundkiewicz2018near,junczys2018approaching,chollampatt2017connecting,ji2017nested}. Currently the most effective approaches regard GEC as machine translation problem that translates erroneous sentences to correct sentences. Thus, many methods that are based on statistical or neural machine translation architectures have been proposed. However, most of the existing GEC systems have focused on correction of grammar errors instead of noised spellings. For example, most of words in a wrong sentence in CoNLL-2014 shared task~\cite{ng2014conll} are correct such as `Nothing is absolute right or wrong', where the only error comes from the specific form `absolute'. One of the existing works that are most similar to this paper is scRNN~\cite{sakaguchi2017robsut}, where each word is represented in a fixed `bag of characters' way. It only consists of a word-level RNN and focused on very easy noise. On the contrary, our proposed framework is more flexible and can obtain meaningful representations that incorporate both character and word-level dependencies. In addition, we have experimented on more difficult types of noise than these in~\cite{sakaguchi2017robsut} and achieved much better performance.
\subsection{Denoising text for downstream tasks}
Robust NLP systems are becoming increasingly important given the severe consequences adversarial samples can cause~\cite{grosse2017adversarial,iyyer2018adversarial,xu2019adversarial}. However, previous works have shown that neural machine translation models can be easily broken with words whose characters are permuted~\cite{belinkov2017synthetic}. To solve this problem, researchers have found that misspelling correction models can play an extremely effective role~\cite{pruthi2019combating,zhou2019improving} in improving the robustness of the systems. For example, Pruthi {\it et al}~\cite{pruthi2019combating} firstly applied the pre-trained scRNN model to source sentence to remove noise and then the denoised source sentence was input into the neural translation model to obtain the correctly translated sentence. In addition, Zhou {\it et al}~\cite{zhou2019improving} directly integrated such denoising models into the machine translation system that was trained in an end-to-end approach. In either way, these works suggest that the proposed framework which has demonstrated strong performance can have great potentials in improving the robustness of other NLP systems.
\section{Conclusion}
As most of the current NLP systems are very brittle, it is extremely important to develop robust neural models. In this paper, we have presented a word recognition framework, MUDE, that achieves very strong and robust performance with different types of noise presenting. The proposed framework is able to capture both character and word-level dependencies to obtain effective word representations. Extensive experiments on datasets with various types of noise have demonstrated its superior performance over the exiting popular models.

There are several meaningful future research directions that are worthy exploring. The first is to extend MUDE to deal with sentences where word-level noise presents. For example, in the noised sentences, some of the words might be swapped, dropped, inserted or replaced, etc. In addition, it is also meaningful to improve the generality of MUDE such that it can achieve strong performance with the presence of various types  of noise not seen in the training dataset. Another possible future direction is to utilize MUDE to improve the robustness other NLP systems including machine translation, reading comprehension, text classification, etc.  Lastly, as this work primarily focuses on English, it would be very meaningful to experiment the proposed framework on other languages.


\bibliographystyle{aaai}
\balance
\bibliography{ref}

\begin{thebibliography}{}

\bibitem[\protect\citeauthoryear{Andermatt, Pezold, and
  Cattin}{2016}]{andermatt2016multi}
Andermatt, S.; Pezold, S.; and Cattin, P.
\newblock 2016.
\newblock Multi-dimensional gated recurrent units for the segmentation of
  biomedical 3d-data.
\newblock In {\em Deep Learning and Data Labeling for Medical Applications}.
  Springer.
\newblock  142--151.

\bibitem[\protect\citeauthoryear{Belinkov and
  Bisk}{2017}]{belinkov2017synthetic}
Belinkov, Y., and Bisk, Y.
\newblock 2017.
\newblock Synthetic and natural noise both break neural machine translation.
\newblock {\em arXiv preprint arXiv:1711.02173}.

\bibitem[\protect\citeauthoryear{Cho \bgroup et al\mbox.\egroup
  }{2014}]{cho2014learning}
Cho, K.; Van~Merri{\"e}nboer, B.; Gulcehre, C.; Bahdanau, D.; Bougares, F.;
  Schwenk, H.; and Bengio, Y.
\newblock 2014.
\newblock Learning phrase representations using rnn encoder-decoder for
  statistical machine translation.
\newblock {\em arXiv:1406.1078}.

\bibitem[\protect\citeauthoryear{Chollampatt and
  Ng}{2017}]{chollampatt2017connecting}
Chollampatt, S., and Ng, H.~T.
\newblock 2017.
\newblock Connecting the dots: Towards human-level grammatical error
  correction.
\newblock In {\em Proceedings of the 12th Workshop on Innovative Use of NLP for
  Building Educational Applications},  327--333.

\bibitem[\protect\citeauthoryear{Davis}{2012}]{davis2012psycholinguistic}
Davis, M.
\newblock 2012.
\newblock Psycholinguistic evidence on scrambled letters in reading.

\bibitem[\protect\citeauthoryear{Devlin \bgroup et al\mbox.\egroup
  }{2018}]{devlin2018bert}
Devlin, J.; Chang, M.-W.; Lee, K.; and Toutanova, K.
\newblock 2018.
\newblock Bert: Pre-training of deep bidirectional transformers for language
  understanding.
\newblock {\em arXiv:1810.04805}.

\bibitem[\protect\citeauthoryear{Dinan \bgroup et al\mbox.\egroup
  }{2019}]{dinan2019build}
Dinan, E.; Humeau, S.; Chintagunta, B.; and Weston, J.
\newblock 2019.
\newblock Build it break it fix it for dialogue safety: Robustness from
  adversarial human attack.
\newblock {\em arXiv:1908.06083}.

\bibitem[\protect\citeauthoryear{Ebrahimi \bgroup et al\mbox.\egroup
  }{2017}]{ebrahimi2017hotflip}
Ebrahimi, J.; Rao, A.; Lowd, D.; and Dou, D.
\newblock 2017.
\newblock Hotflip: White-box adversarial examples for text classification.
\newblock {\em arXiv:1712.06751}.

\bibitem[\protect\citeauthoryear{Fumera, Pillai, and
  Roli}{2006}]{fumera2006spam}
Fumera, G.; Pillai, I.; and Roli, F.
\newblock 2006.
\newblock Spam filtering based on the analysis of text information embedded
  into images.
\newblock {\em Journal of Machine Learning Research} 7(Dec):2699--2720.

\bibitem[\protect\citeauthoryear{Grosse \bgroup et al\mbox.\egroup
  }{2017}]{grosse2017adversarial}
Grosse, K.; Papernot, N.; Manoharan, P.; Backes, M.; and McDaniel, P.
\newblock 2017.
\newblock Adversarial examples for malware detection.
\newblock In {\em European Symposium on Research in Computer Security},
  62--79.
\newblock Springer.

\bibitem[\protect\citeauthoryear{Grundkiewicz and
  Junczys-Dowmunt}{2018}]{grundkiewicz2018near}
Grundkiewicz, R., and Junczys-Dowmunt, M.
\newblock 2018.
\newblock Near human-level performance in grammatical error correction with
  hybrid machine translation.
\newblock {\em arXiv:1804.05945}.

\bibitem[\protect\citeauthoryear{Iyyer \bgroup et al\mbox.\egroup
  }{2018}]{iyyer2018adversarial}
Iyyer, M.; Wieting, J.; Gimpel, K.; and Zettlemoyer, L.
\newblock 2018.
\newblock Adversarial example generation with syntactically controlled
  paraphrase networks.
\newblock {\em arXiv:1804.06059}.

\bibitem[\protect\citeauthoryear{Ji \bgroup et al\mbox.\egroup
  }{2017}]{ji2017nested}
Ji, J.; Wang, Q.; Toutanova, K.; Gong, Y.; Truong, S.; and Gao, J.
\newblock 2017.
\newblock A nested attention neural hybrid model for grammatical error
  correction.
\newblock {\em arXiv:1707.02026}.

\bibitem[\protect\citeauthoryear{Junczys-Dowmunt \bgroup et al\mbox.\egroup
  }{2018}]{junczys2018approaching}
Junczys-Dowmunt, M.; Grundkiewicz, R.; Guha, S.; and Heafield, K.
\newblock 2018.
\newblock Approaching neural grammatical error correction as a low-resource
  machine translation task.
\newblock {\em arXiv:1804.05940}.

\bibitem[\protect\citeauthoryear{Liu \bgroup et al\mbox.\egroup
  }{2019}]{liu2019say}
Liu, H.; Derr, T.; Liu, Z.; and Tang, J.
\newblock 2019.
\newblock Say what i want: Towards the dark side of neural dialogue models.
\newblock {\em arXiv:1909.06044}.

\bibitem[\protect\citeauthoryear{Marcus, Santorini, and
  Marcinkiewicz}{1993}]{marcus1993building}
Marcus, M.; Santorini, B.; and Marcinkiewicz, M.~A.
\newblock 1993.
\newblock Building a large annotated corpus of english: The penn treebank.

\bibitem[\protect\citeauthoryear{Ng \bgroup et al\mbox.\egroup
  }{2014}]{ng2014conll}
Ng, H.~T.; Wu, S.~M.; Briscoe, T.; Hadiwinoto, C.; Susanto, R.~H.; and Bryant,
  C.
\newblock 2014.
\newblock The conll-2014 shared task on grammatical error correction.
\newblock In {\em Proceedings of the Eighteenth Conference on Computational
  Natural Language Learning: Shared Task},  1--14.

\bibitem[\protect\citeauthoryear{Perea \bgroup et al\mbox.\egroup
  }{2015}]{perea2015letter}
Perea, M.; Jim{\'e}nez, M.; Talero, F.; and L{\'o}pez-Ca{\~n}ada, S.
\newblock 2015.
\newblock Letter-case information and the identification of brand names.
\newblock {\em British Journal of Psychology} 106(1):162--173.

\bibitem[\protect\citeauthoryear{Pruthi, Dhingra, and
  Lipton}{2019}]{pruthi2019combating}
Pruthi, D.; Dhingra, B.; and Lipton, Z.~C.
\newblock 2019.
\newblock Combating adversarial misspellings with robust word recognition.
\newblock {\em arXiv:1905.11268}.

\bibitem[\protect\citeauthoryear{Rayner, White, and
  Liversedge}{2006}]{rayner2006raeding}
Rayner, K.; White, S.~J.; and Liversedge, S.
\newblock 2006.
\newblock Raeding wrods with jubmled lettres: There is a cost.

\bibitem[\protect\citeauthoryear{Sakaguchi \bgroup et al\mbox.\egroup
  }{2017}]{sakaguchi2017robsut}
Sakaguchi, K.; Duh, K.; Post, M.; and Van~Durme, B.
\newblock 2017.
\newblock Robsut wrod reocginiton via semi-character recurrent neural network.
\newblock In {\em Thirty-First AAAI Conference on Artificial Intelligence}.

\bibitem[\protect\citeauthoryear{Sutskever, Martens, and
  Hinton}{2011}]{sutskever2011generating}
Sutskever, I.; Martens, J.; and Hinton, G.~E.
\newblock 2011.
\newblock Generating text with recurrent neural networks.
\newblock In {\em ICML-11},  1017--1024.

\bibitem[\protect\citeauthoryear{Tieleman and
  Hinton}{2012}]{tieleman2012rmsprop}
Tieleman, T., and Hinton, G.
\newblock 2012.
\newblock Rmsprop.
\newblock {\em COURSERA: Lecture} 7017.

\bibitem[\protect\citeauthoryear{Vaswani \bgroup et al\mbox.\egroup
  }{2017}]{vaswani2017attention}
Vaswani, A.; Shazeer, N.; Parmar, N.; Uszkoreit, J.; Jones, L.; Gomez, A.~N.;
  Kaiser, {\L}.; and Polosukhin, I.
\newblock 2017.
\newblock Attention is all you need.
\newblock In {\em Advances in neural information processing systems},
  5998--6008.

\bibitem[\protect\citeauthoryear{Wolf, Miller, and
  Grodzinsky}{2017}]{wolf2017we}
Wolf, M.~J.; Miller, K.; and Grodzinsky, F.~S.
\newblock 2017.
\newblock Why we should have seen that coming: comments on microsoft's tay
  experiment, and wider implications.
\newblock {\em ACM SIGCAS Computers and Society} 47(3):54--64.

\bibitem[\protect\citeauthoryear{Xu \bgroup et al\mbox.\egroup
  }{2019a}]{xu2019adaptive}
Xu, D.; Cheng, W.; Luo, D.; Gu, Y.; Liu, X.; Ni, J.; Zong, B.; Chen, H.; and
  Zhang, X.
\newblock 2019a.
\newblock Adaptive neural network for node classification in dynamic networks.
\newblock In {\em ICDM}.
\newblock IEEE.

\bibitem[\protect\citeauthoryear{Xu \bgroup et al\mbox.\egroup
  }{2019b}]{xu2019spatio}
Xu, D.; Cheng, W.; Luo, D.; Liu, X.; and Zhang, X.
\newblock 2019b.
\newblock Spatio-temporal attentive rnn for node classification in temporal
  attributed graphs.
\newblock In {\em IJCAI},  3947--3953.

\bibitem[\protect\citeauthoryear{Xu \bgroup et al\mbox.\egroup
  }{2019c}]{xu2019adversarial}
Xu, H.; Ma, Y.; Liu, H.; Deb, D.; Liu, H.; Tang, J.; and Jain, A.
\newblock 2019c.
\newblock Adversarial attacks and defenses in images, graphs and text: A
  review.
\newblock {\em arXiv:1909.08072}.

\bibitem[\protect\citeauthoryear{Yang \bgroup et al\mbox.\egroup
  }{2019}]{yang2019xlnet}
Yang, Z.; Dai, Z.; Yang, Y.; Carbonell, J.; Salakhutdinov, R.; and Le, Q.~V.
\newblock 2019.
\newblock Xlnet: Generalized autoregressive pretraining for language
  understanding.
\newblock {\em arXiv preprint arXiv:1906.08237}.

\bibitem[\protect\citeauthoryear{Yu \bgroup et al\mbox.\egroup
  }{2018}]{yu2018qanet}
Yu, A.~W.; Dohan, D.; Luong, M.-T.; Zhao, R.; Chen, K.; Norouzi, M.; and Le,
  Q.~V.
\newblock 2018.
\newblock Qanet: Combining local convolution with global self-attention for
  reading comprehension.
\newblock {\em arXiv:1804.09541}.

\bibitem[\protect\citeauthoryear{Zhao \bgroup et al\mbox.\egroup
  }{2019}]{zhao2019improving}
Zhao, W.; Wang, L.; Shen, K.; Jia, R.; and Liu, J.
\newblock 2019.
\newblock Improving grammatical error correction via pre-training a
  copy-augmented architecture with unlabeled data.
\newblock {\em arXiv:1903.00138}.

\bibitem[\protect\citeauthoryear{Zhao, Dua, and
  Singh}{2017}]{zhao2017generating}
Zhao, Z.; Dua, D.; and Singh, S.
\newblock 2017.
\newblock Generating natural adversarial examples.
\newblock {\em arXiv:1710.11342}.

\bibitem[\protect\citeauthoryear{Zhou \bgroup et al\mbox.\egroup
  }{2019}]{zhou2019improving}
Zhou, S.; Zeng, X.; Zhou, Y.; Anastasopoulos, A.; and Neubig, G.
\newblock 2019.
\newblock Improving robustness of neural machine translation with multi-task
  learning.
\newblock In {\em Proceedings of the Fourth Conference on Machine Translation
  (Volume 2: Shared Task Papers, Day 1)},  565--571.

\end{thebibliography}
\end{document}